# Incorporating uncertainty quantification into travel mode choice modeling: a Bayesian neural network (BNN) approach and an uncertainty-guided active survey framework

Shuwen Zheng, Zhou Fang[*], Liang Zhao[*]
*School of Architecture, Tsinghua University, Beijing, 100084, China*

**Abstract:**
Existing deep learning approaches for travel mode choice modeling fail to inform modelers about their prediction uncertainty. Even when facing scenarios that are out of the distribution of training data, which implies high prediction uncertainty, these approaches still provide deterministic answers, potentially leading to misguidance. To address this limitation, this study introduces the concept of uncertainty from the field of explainable artificial intelligence into travel mode choice modeling. We propose a Bayesian neural network-based travel mode prediction model (BTMP) that quantifies the uncertainty of travel mode predictions, enabling the model itself to "know" and "tell" what it doesn't know. With BTMP, we further propose an uncertainty-guided active survey framework, which dynamically formulates survey questions representing travel mode choice scenarios with high prediction uncertainty. Through iterative collection of responses to these dynamically tailored survey questions, BTMP is iteratively trained to achieve the desired accuracy faster with fewer questions, thereby reducing survey costs. Experimental validation using synthetic datasets confirms the effectiveness of BTMP in quantifying prediction uncertainty. Furthermore, experiments, utilizing both synthetic and real-world data, demonstrate that the BTMP model, trained with the uncertainty-guided active survey framework, requires 20% to 50% fewer survey responses to match the performance of the model trained on randomly collected survey data. Overall, the proposed BTMP model and active survey framework innovatively incorporate uncertainty quantification into travel mode choice modeling, providing model users with essential insights into prediction reliability while optimizing data collection for deep learning model training in a cost-efficient manner.

**Highlights:**
- Incorporating uncertainty quantification into travel mode choice modeling.
- Proposing a BNN-based model that quantifies the uncertainty of travel mode predictions.
- Proposing an uncertainty-guided active survey framework that optimizes the data collection process for the cost-effective development of the deep learning based travel mode choice prediction model.
- Designing two experiments using synthetic and real-world data to examine the effectiveness of the proposed model and framework.

---

[*] Corresponding author.
*E-mail address:* fangzhou@mail.tsinghua.edu.cn (Zhou Fang), zhaoliang@tsinghua.edu.cn (Liang Zhao).

# 1 Introduction

Modeling travel mode choice is a critical component within the classic four-stage transport planning process, and is essential for predicting travel demand and understanding its influencing factors (de Dios Ortúzar and Willumsen, 2011). With the advancement of artificial intelligence, deep learning based models are increasingly being developed and applied in the travel mode choice modeling due to their high approximation power and versatility in handling diverse data types.

Unlike the traditionally widely applied discrete choice models (DCM) for travel mode choice prediction (Ben-Akiva and Lerman, 1985), which employ random utility theory as a predefined universal decision rule (Train, 2003), deep learning approaches need to derive all model weights from training data to form decision boundaries for mode choice predictions. This highlights two critical issues in developing and employing deep learning models in the field.

First, in high-risk fields of urban and transport planning, modellers not only prioritize accuracy in mode choice prediction results but also require access to the model's confidence levels in their predictions. For deep learning models, when making predictions for scenarios beyond the scope of its training data, such as new unseen combinations of travel time and cost of various mode alternatives, the model may exhibit high prediction uncertainty due to lack of experience and difficulty in generalizing to those contexts. In such cases, model users need to be informed of this information and question or modify model prediction outcomes promptly to alleviate the risk of potential misguidance. Therefore, it's necessary for deep learning models to provide information about whether the model trained on existing data can generalize to predict new scenarios, i.e. whether the model has seen similar data combinations and can make confidence predictions. This poses a significant need to quantify the uncertainty when the model predicts specific scenarios.

Second, it's also critical to efficiently acquire sufficient representative training data for deep learning models to build up generalizable prediction capabilities while minimizing the survey cost. Studies have demonstrated that training neural networks requires a larger amount of data compared to traditional DCM approach. A sufficiently large sample size (on the order of $10^4$) may be necessary to ensure the superiority of deep learning model performances over traditional DCM (Wang et al., 2021b). This emphasizes the importance of providing sufficient training data from travel mode choice surveys for the development of deep learning models. However, travel mode choice surveys often require significant time and effort. Overly extensive and comprehensive travel mode choice surveys not only demand considerable effort from researchers, but may also lead respondents to provide inaccurate or ill-considered answers due to fatigue effect (Bateman et al., 2008; Holmes and Boyle, 2005). Therefore, in order to balance the predictive performances of deep learning models and the budget for conducting survey, there is an obvious need to inquire about the most informative questions for the model during the survey process, allowing the model to reach higher accuracy with minimized number of questions asked, thereby reducing the survey cost.

However, current deep learning models in travel mode choice modeling fail to address the two critical issues mentioned above. First, the "end-to-end" structure of existing deep learning models in travel mode choice modeling can only provide a single prediction outcome, without providing users with information about the uncertainty of the model prediction. Researchers are not informed about where the model already has enough data, and where the model needs more data (i.e. the model is less confident about its predictions here due to the lack of specific types of training data). Second, existing studies that apply deep learning models in travel mode choice mostly focus on model fitting performance and prediction accuracy based on given datasets (e.g., Martín-Baos et al., 2023; Wang et al., 2020; Xia et al., 2023). Limited attention has been given to proposing a pipeline that initiates from data collection stage and focuses on collecting high-quality data in a more cost-effective manner.

To address the two critical issues in the field and the limitations of existing deep learning approaches, this study introduces the concept of uncertainty from the field of explainable artificial intelligence into travel mode choice modeling. We first proposes a Bayesian neural network (BNN) based travel mode prediction model, BTMP, which is equipped with the capability to quantify the model prediction uncertainty through MC Dropout technique. Leveraging the BTMP's capability to quantify uncertainty of model predictions, we further develop an uncertainty-guided active survey framework, which dynamically formulates survey questions representing travel mode choice scenarios with high prediction uncertainty identified by gradually trained BTMP model. Through iterative collection of responses to these dynamically tailored survey questions, BTMP model progressively gains informative training data and builds up its predictive capability through model training. Meanwhile, the iteratively trained BTMP model can continuously guide active survey process by formulating new questions with updated uncertainty information. Two experiments, using synthetic and real-world data, were designed and delivered to examine the effectiveness of the proposed BTMP model and active survey framework.

The primary contributions of the paper can be summarized as follows. First, this research incorporates uncertainty quantification into travel mode choice modeling by proposing a BNN-based travel mode prediction model, BTMP, allowing users to obtain not only the prediction results but also the model uncertainty of these predictions. Second, this research further proposes a framework for active survey in travel mode choice modeling guided by uncertainty quantification, which enables more effective data acquisition and cost savings in surveys.

The remainder of this paper is organized as follows. Section 2 presents a literature review on deep learning approaches in travel mode choice modeling, BNN and uncertainty in the field of explainable AI, and the applications of BNN in the field of transportation. Section 3 illustrates the proposed BNN-based travel mode prediction model (BTMP) and the proposed active survey framework for travel mode choice modeling. Section 4 describes the design and the results of the experiments used for demonstrating the effectiveness of the proposed model and framework. Section 5 concludes the paper with remarks on its contribution.

## 2 Literature Review
### 2.1 Deep learning approaches in travel mode choice modeling

With the advancement of artificial intelligence technology, deep learning methods have been increasingly applied in travel mode choice modeling, serving as an alternative to traditional logit models. Nijkamp et al. (1996) pioneered the introduction of deep learning techniques in travel mode choice modeling, exploring the mode choice patterns between railway and road transport modes in Italy. Subsequently, Subba Rao et al. (1998), Xie et al. (2003), Cantarella and de Luca (2005) also explored the application of neural networks in travel mode choice modeling. In recent years, a series of studies have focused on comprehensive comparisons between discrete choice models, neural networks, and other existing machine learning models, further demonstrating that deep learning approaches exhibit superior predictive performance compared to traditional discrete choice models (Hagenauer and Helbich, 2017; Karlaftis and Vlahogianni, 2011; Martín-Baos et al., 2023; Wang et al., 2021a; Zhao et al., 2020).

However, existing research on deep learning models for travel mode choice exhibits two research gaps. Firstly, current deep learning models for travel mode choice typically employ an "end-to-end" training approach, resulting in a single prediction outcome without providing information about the uncertainty of the model predictions, which can arise due to the complexity of the underlying relationships in the training data, or the lack of specific types of training data. The inability to quantify uncertainty prevents researchers from knowing where the model lacks confidence in its predictions, which can lead to misleading decisions. It also fails to inform researchers about when they should pay more attention during model development, in other words, where the model requires more training data to achieve better predictive performances.

Secondly, while existing studies on deep learning models for travel mode choice prediction primarily focus on enhancing prediction accuracy on given datasets (Martín-Baos et al., 2023; Wang et al., 2020; Xia et al., 2023), limited attention has been paid to developing a tailored survey process for deep learning approaches to acquire high-quality training data. Unlike rule-based discrete choice models, which employ random utility theory as a predefined universal decision rule, deep learning approaches need to derive all model weights from training data to form decision boundaries for mode choice predictions. Compared to DCM approach, the predictive performance and generalization capability of deep learning models rely significantly more on both the quantity and quality of data (Wang et al., 2021b). Considering that acquiring these training data from travel mode choice surveys can be costly and labor-intensive for researchers, there is a pressing need to obtain sufficient and informative training data for the model development within the survey budget. However, none of the existing studies has proposed a framework starting from the data collection stage that can effectively collect the most informative data with minimized questions asked, and achieve better model performances for travel mode choice modeling in a cost-efficient manner.

**2.2 Bayesian neural networks and uncertainty in the field of explainable AI**
In the field of machine learning and explainable AI, exploring the uncertainty of model predictions is also a critical issue (Ghahramani, 2015). In many high-risk fields with high-stakes decisions, such as autonomous driving and medical diagnostics, it is crucial for model users to know the model's confidence levels in their predictions and not to trust them blindly. Uncertainty of model predictions can help users ensure that the model is working as expected, allowing the decision-makers to question or modify model prediction outcomes to alleviate the potential risk when the model exhibits high uncertainty.

Among all the methods to incorporate uncertainty quantification into deep learning models, Bayesian neural networks (BNNs) provide a practical framework for understanding uncertainty, and have been widely applied in numerous fields, including autonomous driving (Kendall, 2017; Ravindran et al., 2022), medical image analysis (Chen et al., 2022) and security monitoring (Zafar et al., 2019). BNN, first introduced by Mackay (1992), replaces the model's weights from single values to distributions (Blundell et al., 2015), which results in non-fixed and probabilistic model predictions, incorporating the concept of uncertainty into conventional deep learning models. To implement BNN and perform model predictions, Monte Carlo dropout (MC Dropout) is a widely adopted and mainstream method (Gal, 2016; Gal and Ghahramani, 2015; Kendall and Gal, 2017; Kendall, 2017). This technique involves training a model with random dropout applied before each weight layer, and also applying dropout when conducting model predictions. Adopting MC Dropout technique, Kendall and Gal (2017) further proposed the method of quantifying uncertainty (also referred to as epistemic uncertainty) in classification problems by calculating the entropy of the prediction results.

The proven capability of BNN to quantify model prediction uncertainty provides us with an opportunity to introduce explainable AI and uncertainty into the field of travel mode choice, allowing us to address the limitations of existing deep learning models in travel mode choice modeling.

**2.3 The applications of Bayesian neural networks in the field of transportation**
Equipped with the capability to quantify uncertainty, BNN has been widely applied and popular in numerous high-risk fields. However, there are only a few articles that have utilized BNNs in the field of transportation, which is also a high-risk field with high-stakes decisions, and the limited attempts of using BNN in transportation studies are far from fully utilizing the capability of BNN to quantify uncertainty. Cui et al. (2018) employed BNN to infer trip purpose with social media data and POIs, Jia et al. (2020) and Xia et al. (2023) adopted BNN to improve accuracy in travel mode choice predictions. However, none of the existing studies utilizes BNN to quantify the uncertainty of model predictions and provide guidance on acquiring survey data more efficiently. In this context, the ability of BNN to quantify uncertainty of model predictions provides a golden opportunity to address the limitations of existing deep learning models mentioned in Section 2.1.

**3 Methodology**
This study introduces a BNN-based travel mode prediction (BTMP) model, which allows researchers to obtain not only the prediction results from BTMP but also the uncertainty of these predictions. Furthermore, we propose an uncertainty-guided active survey framework with BTMP embedded, which dynamically formulates survey questions representing travel mode choice scenarios with high prediction uncertainty identified by gradually trained BTMP. Through iterative collection of responses to these dynamically tailored survey questions, BTMP gains informative training data, and achieves the desired accuracy faster with fewer questions, thereby reducing survey costs. In this section, we will first introduce the BTMP model, which forms the basic unit of this framework, and then we will give a comprehensive description of the entire active survey framework.

**3.1 BNN-based travel mode prediction (BTMP) model**
Our fundamental model, namely BNN-based travel mode prediction (BTMP) model, builds up its mode choice prediction capabilities through model training on labeled training data pairs, and is equipped with the capability to quantify prediction uncertainty through MC Dropout technique.

**Model structure**: The BTMP model takes the attributes of mode choice alternatives as input (e.g., travel time and cost). The input data is then fed into two hidden layers, each with 128 neurons. Dropout is applied to the two hidden layers with probabilities of 0.25 and 0.5 on the first and second hidden layers respectively to randomly deactivate some neurons. Finally, a softmax function converts the data to a series of values in the [0,1] range, which represents the probability distribution of choosing each mode choice alternative. By comparing the alternative with the highest probability, the model predicts the chosen travel mode alternative as its output.

**Loss function**: In this research, Cross-entropy loss is minimized to train the proposed BTMP model by iteratively updating the model parameters. Cross-entropy is a widely used loss function in classification problems such as travel mode choice. It builds upon the idea of information entropy theory and measures the difference between two probability distributions for a given random variable/set of events. In the proposed BTMP model, we adopt Cross-entropy loss to quantify the information difference between the predicted probability distribution (output from the models) and the ground truth probability distribution (travel mode choice result obtained from survey) following the equation:

$$E(p,q) = -\sum_{x \in X} p(x) \log q(x) \tag{1}$$

where $p(x)$ is the ground truth distribution and $q(x)$ is the predicted distribution from the models.

**Model training process**: The attributes of mode choice alternatives form the model input (e.g., travel time and cost), and the travel mode choice result obtained from survey form the ground truth labels. The model first gets predictions for the travel mode choice given the input features, then calculates the Cross-entropy loss for model training using **Eq. (1)** and updates the model parameters with the calculated loss through backpropagation process. Using the updated parameters, the model gets predictions and calculates the loss again. This iterative process continues until the Cross-entropy loss for models on the training dataset maintains a stable level, and the model training process is completed.

**3.2 The application of BTMP model to quantify uncertainty of model predictions**

The process of training the model and using MC Dropout to quantify uncertainty of model predictions is illustrated in **Fig. 1**. In stage 1, the BTMP model is trained with labeled training data pairs, which contain the attributes of mode choice alternatives as input features, and mode choices obtained from survey as corresponding ground truth labels. After the training process has been completed, we obtain the trained BTMP model with calibrated model parameters. In Stage 2, while keeping the model parameters fixed, we enable the random dropout of the trained BTMP. Then the unlabeled data $q$, which only contains input features with unknown ground truth travel mode choice, is fed into the trained BTMP for uncertainty quantifying, and we conduct model predictions with random dropout for $T$ times. After the prediction process has finished, we calculate the entropy of probability of choosing each mode $c$, as the uncertainty of model predictions for question $q$ using the following two equations (Kendall and Gal, 2017):

$$\overline{p_{q,c}} = \frac{1}{T}\sum_{t=1}^{T} p_{q,c}(\mathbf{x}_q, \hat{\mathbf{w}}_t) \qquad (2)$$

$$U_q = -\sum_{c=1}^{C} \overline{p_{q,c}} \log \overline{p_{q,c}} \qquad (3)$$

where $\mathbf{x}_q$ is the input features of question $q$, and $\hat{\mathbf{w}}_t$ is the sampled model parameters through random dropout in iteration $t$.

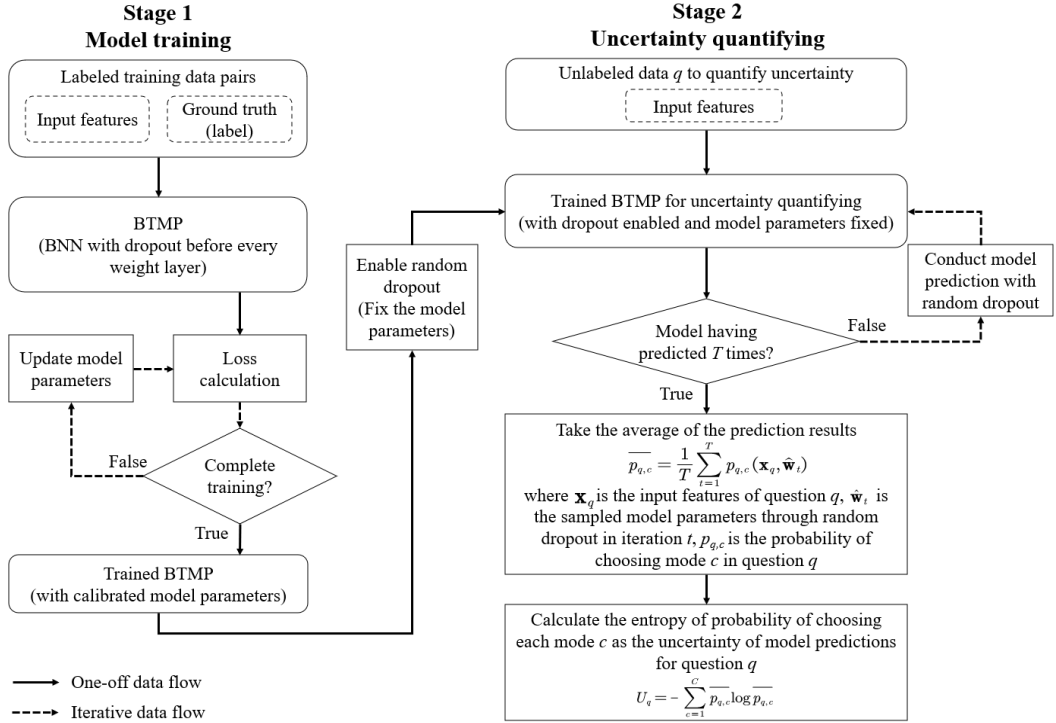

**Fig. 1. The training and uncertainty quantifying process of the BTMP model**

**3.3 Active survey framework in travel mode choice prediction with BTMP embedded**

The BTMP model and its capability to quantify uncertainty provide a golden opportunity to implement active survey strategies in travel mode choice prediction, where the model dynamically formulates survey questions representing travel mode choice scenarios with high prediction uncertainty identified by gradually trained BTMP. Researchers can then iteratively collect responses to these dynamically tailored survey questions, allowing the model to achieve the desired accuracy faster with fewer questions, thereby reducing research costs.

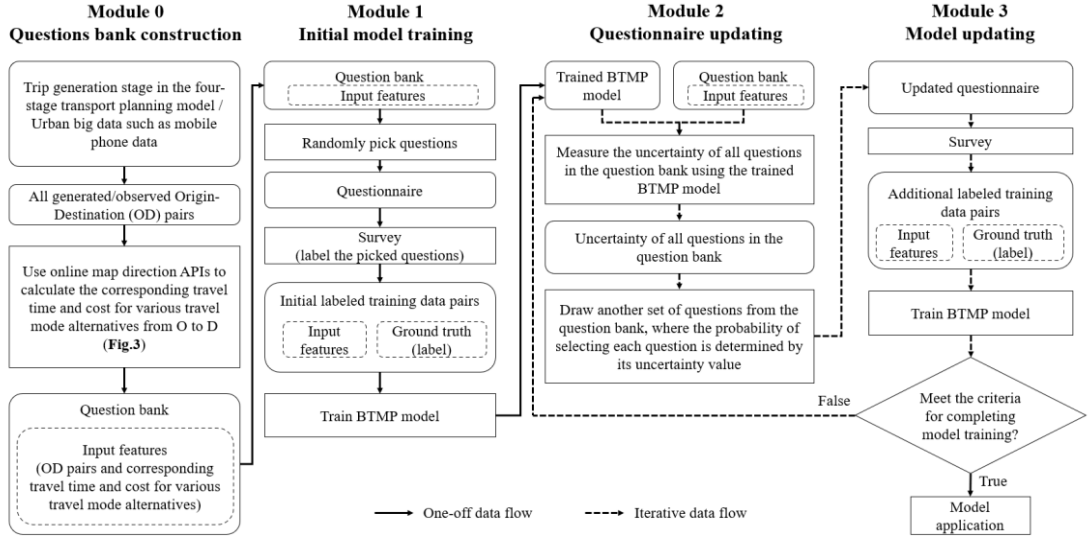

**Fig. 2. Modules and workflow of the proposed framework for active survey**

**Fig. 2** shows the modules and workflow of the proposed framework for the uncertainty-guided active survey. In Module 0, we first construct a question bank containing all potential survey questions, with all questions inquiring about the respondents' travel mode choice for various Origin-Destination (OD) pairs. In practice, these OD pairs can be derived from the trip generation stage in the traditional four-stage transport planning model, or be observed from urban big data such as mobile phone data. Then by requesting directions with matching Origin (O) and Destination (D) as the observed OD pairs for all travel modes using online map direction APIs, the corresponding travel time and cost for various travel mode alternatives from O to D are calculated (Hillel, 2019). This process is illustrated with an example of the travel time and cost generated by the online map direction API for an example OD pair in **Fig. 3**.

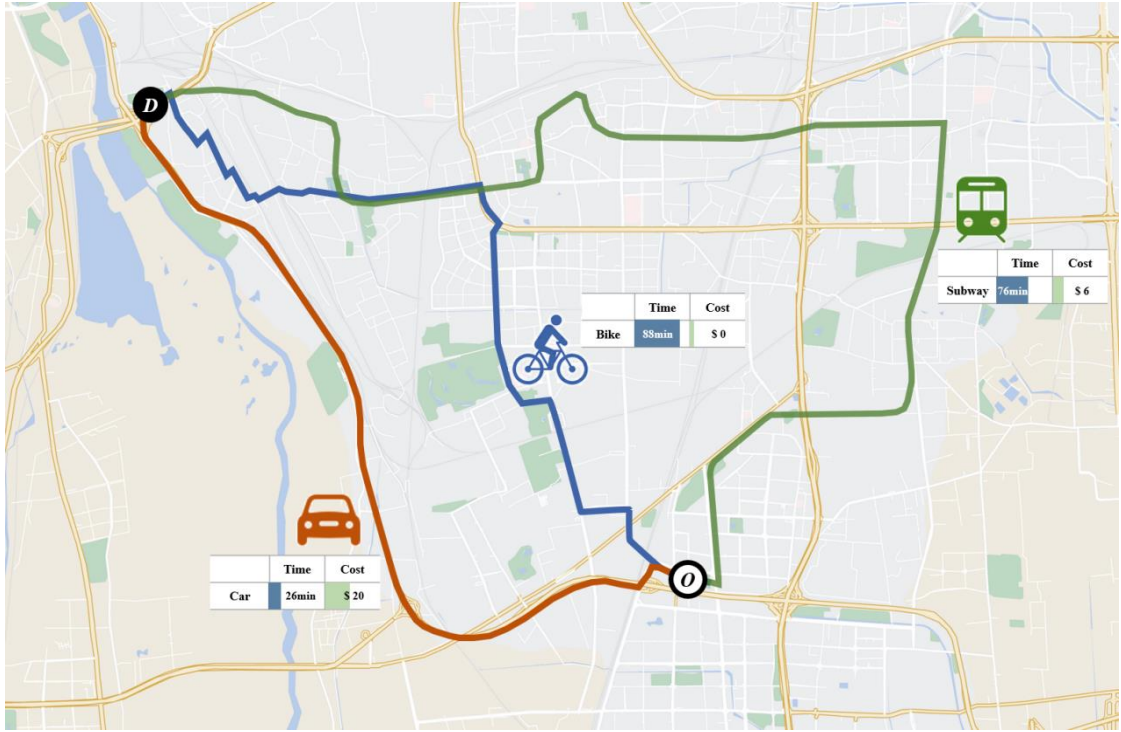

**Fig. 3**. **Travel time and cost calculated by the online map direction API for an example OD pair**

By combining the OD pairs and the calculated travel time and cost of their corresponding mode alternatives, the question bank is developed and serves as the input features of the training dataset (Hillel et al., 2018). Researchers then randomly select a subset of questions from the question bank to form the questionnaire, distribute it, and collect the travel mode choice responses, which serve as the ground truth of the initial training dataset. After combining the input features with the corresponding ground truths, we create the initial training dataset, which is then used to train the BTMP model.

In Module 2, all questions in the question bank containing only input features with unknown ground truth mode choices are fed into the initially trained BTMP model. The uncertainties of all questions are then measured using the MC Dropout technique. The uncertainties of model predictions on different questions represent the importance of different questions for model training, informing researchers where the model already has sufficient data (i.e., low uncertainty) and where the model requires more data (i.e., high uncertainty). Researchers are then able to update the questionnaire by drawing another set of questions from the question bank that represent travel mode choice scenarios with high prediction uncertainty quantified by BTMP. Each question $q$ will be selected to form the updated questionnaire with a probability of $p_q$, which is calculated using the quantified uncertainty of this question $U_q$:

$$p_q = \frac{U_q}{\sum_q U_q} \qquad (3)$$

After the questionnaire has been updated with another set of questions based on their uncertainty values, in Module 3, the updated questionnaire is distributed, and the newly collected data will be supplemented to the previously obtained dataset, forming the training data for the new iteration. Subsequently, a new model is trained using the updated training dataset, and the newly trained model is then used in Module 2 to measure the uncertainty of model predictions for each question in the question bank and draw another set of questions according to the acquisition function. This iterative process continues, gradually enlarging the training set, until the model meets the criteria for completing model training and the iteration stops. We define two criteria for completing model training: (1) the model's prediction accuracy does not significantly increase with additional data supplementation for a certain number of iterations; (2) survey costs have reached or exceeded the budget, or for other reasons, it is not feasible to distribute any more questionnaires. Meeting either of these criteria stops the active survey iteration and signals the completion of training, allowing for the application of the trained model.

## 4 Experiments
### 4.1 Preliminaries

To test the performance of BTMP model and demonstrate the effectiveness of the proposed uncertainty-guided active survey framework, we design an iterative training data sampling framework as an alternative to simulate the active survey process. First, we generate all questions in the question bank, with each question including an OD pair and the travel time and cost of mode choice alternatives. Then we assign ground truth labels to all the questions, which are the travel mode choice results for all the OD pairs. All these questions and ground truth labels constitute a data pool used for model training, validation, and testing. We assume that BTMP model does not have access to all ground truth labels at the beginning. Instead, it can progressively obtain the ground truth labels that correspond to the iteratively selected additional training data from the data pool.

The alternative framework of the active survey process is illustrated in detail in **Fig. 4**. In Module 1, we conduct simple random sampling from the data pool across different Destination (D) places to obtain initial training data. This dataset contains mode choice alternative attributes as input features and corresponding ground truth mode choices, and is used for training the BTMP model. In Module 2, the travel time and cost of all trip cases in the data pool are fed into the trained BTMP model, and the average model prediction uncertainty for trips to each destination is measured. This framework then samples additional training data from the data pool using different sampling strategies. In Module 3, sampled additional training data with ground truth labels are added to the existing training dataset for BTMP model training. The trained model is further employed to quantify uncertainty for various destinations and iteratively acquire the additional training data, and the iteration continues until the model's prediction accuracy does not significantly increase with additional data supplementation for a certain number of iterations.

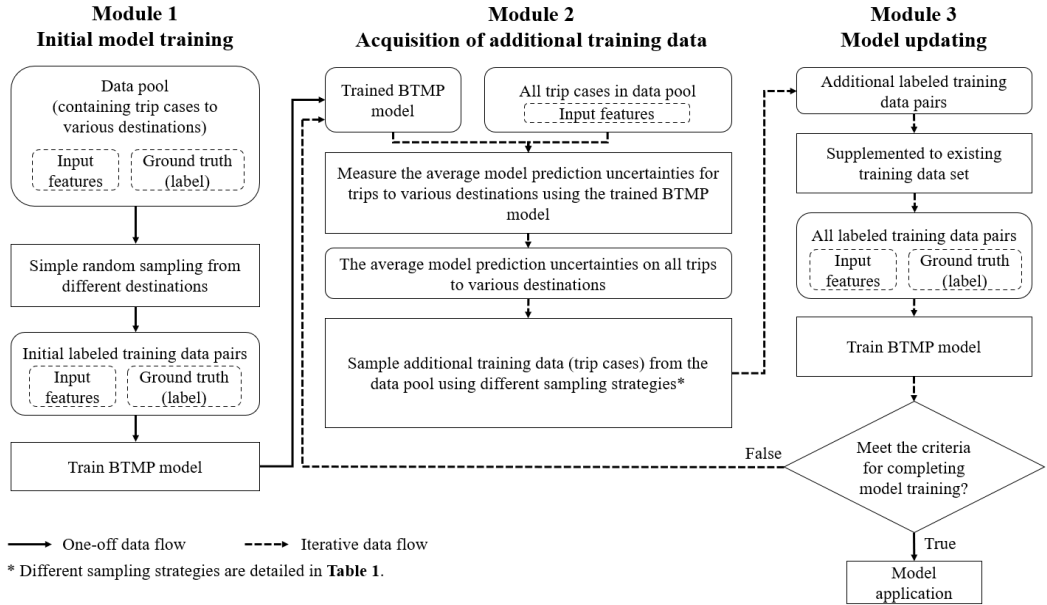

Fig. 4. Alternative framework of the active survey process

**4.2 Experiment design**

Adopting the alternative framework of the active survey process, we design two experiments to examine the BTMP model and the proposed uncertainty-guided active survey framework. Experiment 1 aims to demonstrate the capability of the BTMP model in quantifying uncertainties of model predictions. The set of Experiments 2 aims to demonstrate the effectiveness of the active survey framework for travel mode choice modeling in saving training data samples and reducing survey costs, with synthetic dataset used in Experiment 2-1 and real-case dataset used in Experiment 2-2.

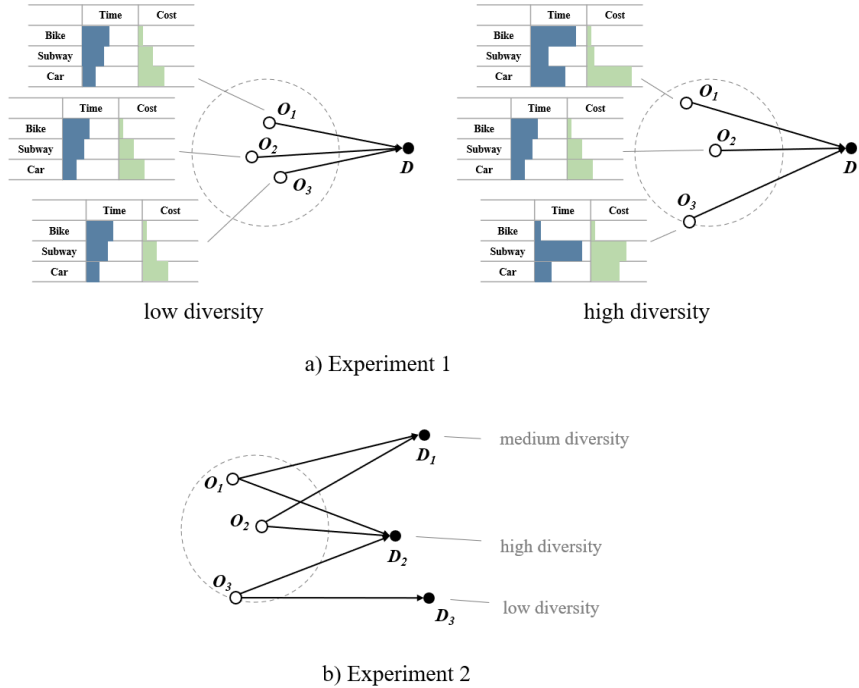

Fig. 5. Different scenarios simulated by theoretical experiments

In Experiment 1, we construct four synthetic datasets, each containing 10,000 OD pairs to one single destination. We assign a set of generated travel time and cost as input features of alternative travel modes for each OD pair within the four datasets. By controlling the diversity of the input features within in each dataset, we obtain four different levels of the similarity of training and testing data samples within the four synthetic datasets for BTMP model development. In reality, these represent the differences in the diversity of the spatial locations of origin places, and the diversity in transportation service levels of various origins among the four dataset (diagrammatic illustration shown in Fig 5a). A dataset with low diversity exhibits higher similarity and closer distributions among training and testing data, making it easier for the model to generalize to the test set. Conversely, a dataset with high diversity shows greater disparities in the distribution between training and testing data, which leads to a higher probability of encountering unseen scenarios in the test set and makes it more difficult for the model to generalize to the test set. We train and test the BTMP model with four datasets respectively, and reveal that the levels of quantified uncertainty across the four datasets align with the levels of their diversity. This aims to demonstrate the capability of the BTMP model in quantifying model prediction uncertainties and to prove the feasibility of utilizing model prediction uncertainty to guide cost-efficient data collection.

Furthermore, to demonstrate the effectiveness of the uncertainty-guided active survey framework in saving training data samples and reducing survey costs, we design the set of Experiments 2 which contains two sub-experiments, named 2-1 and 2-2. Experiment 2-1 constructs a synthetic dataset that contains 10,000 trip cases from multiple Origin (O) places to multiple Destination (D) places, with various diversity levels of data samples for different destinations. Experiment 2-2 utilizes real travel mode choice data obtained from an extensive survey at a typical university in Beijing, China. Both Experiment 2-1 and Experiment 2-2 adopt the alternative framework of the proposed active survey process to develop BTMP models. We compare the BTMP model's predictive performance and the amount of required training data when using the proposed uncertainty-guided sampling strategy (Method 3) versus benchmark sampling strategies (Method 1 and 2), and demonstrate the effectiveness of the proposed active survey framework in saving samples and reducing survey costs. Method 1 employs simple random sampling from different destinations to collect additional training data samples to train BTMP model. Method 2 employs random sampling from the most frequently visited destinations. Method 3 determines sampling probabilities for different destinations based on the quantified uncertainty of various destinations, which is the sampling strategy used in the proposed uncertainty-guided active survey framework.

**Table 1**
**Different sampling strategies of iteratively acquiring additional training data**

|  | Experiment1 | Experiment2-1 | Experiment2-2 |
| --- | --- | --- | --- |
| Method 1 (Simple random sampling from different destinations) | √ | √ | √ |
| Method 2 (Random sampling from most frequently visited destinations) |  |  | √ |
| Method 3 (Sampling probabilities for different destinations associated with uncertainty, **Eq.(3)**) |  | √ | √ |

### 4.3 Experiment 1
**Dataset preparation**

In Experiment 1, we generate four different synthetic datasets, each containing 10,000 OD pairs to the same destination (**Fig. 5a**). These four datasets, named Dataset 1,2,3 and 4 are constructed following the process introduced in **Appendix A** and contain training and testing data samples for model development in four different levels of diversity. Dataset 1 (low diversity) exhibits higher similarity and closer distributions between training and testing data, making it easier for the model to generalize to the test set. Conversely, Dataset 4 (high diversity) shows greater disparities in the distribution between training and testing data, which leads to a higher probability of encountering unseen scenarios in the test set and makes it more difficult for the model to generalize to the test set.

For Dataset 1 to 4, we generate the travel mode choice results among bike, subway, and car based on random utility theory and generated travel time and cost for alternative travel modes, to simulate the decision rules that humans follow (see **Appendix B** for details). For each dataset, the travel time and cost for biking, subway and driving serve as input features, and the travel mode choice results among bike, subway, and car serve as the ground truths. These input features and ground truths constitute all the data pairs (trip cases) used for BTMP model training, validation, and testing. For each dataset, we split 30% of the trip cases as the test set. From the remaining data, we randomly select 200 cases as the validation set, 60 cases as the initial training set, and other trip cases remain in the data pool for uncertainty quantification and additional data acquisition.

**Experiment settings**

We conduct the experiment using Dataset 1, 2, 3 and 4 respectively following the alternative framework of the active survey process proposed in Section 4.1 and **Fig. 4**. For each dataset, we randomly select 60 cases as the initial training set, and acquire 60 trip cases in each iteration as additional training data through simple random sampling from the data pool. In each iteration, we use MC Dropout technique to quantify the uncertainty of the model predictions on test set following the process mentioned in Section 3.2 and **Fig. 1**. We observe the uncertainty and accuracy of model predictions on test set in each iteration for BTMP training using Datasets 1, 2, 3, and 4, respectively. The experiment is repeated five times and the results are averaged.

**Experiment results**

As shown in **Fig. 6**, using the same amount of training data, BTMP model trained on dataset with higher diversity exhibits higher quantified model prediction uncertainty. Additionally, as training data is iteratively supplemented, the uncertainty gradually decreases for all cases. These results demonstrate the BTMP model's capability to quantify uncertainty of model predictions, and this uncertainty can be explained away by supplementing enough training data, aligning with the definition of epistemic uncertainty by Kendall and Gal (2017).

Moreover, when achieving the same level of accuracy, the amount of training data required for Datasets 1, 2, 3, and 4 gradually increases. The results demonstrate that a dataset with higher diversity exhibits greater differences in distributions between training and testing data, leading to a higher probability of encountering unseen scenarios in the test set, making it more challenging, or in other words, requiring more training samples for the model to generalize to the test set. This provides us with an opportunity to utilize the capability of BTMP to quantify uncertainty to indicate where the model needs more data for generalization by identifying OD pairs with higher diversity. Researchers can then iteratively collect the travel mode choice responses to the dynamically tailored questions about these OD pairs, allowing the model to achieve the desired accuracy faster with selective questions, thereby reducing research costs.

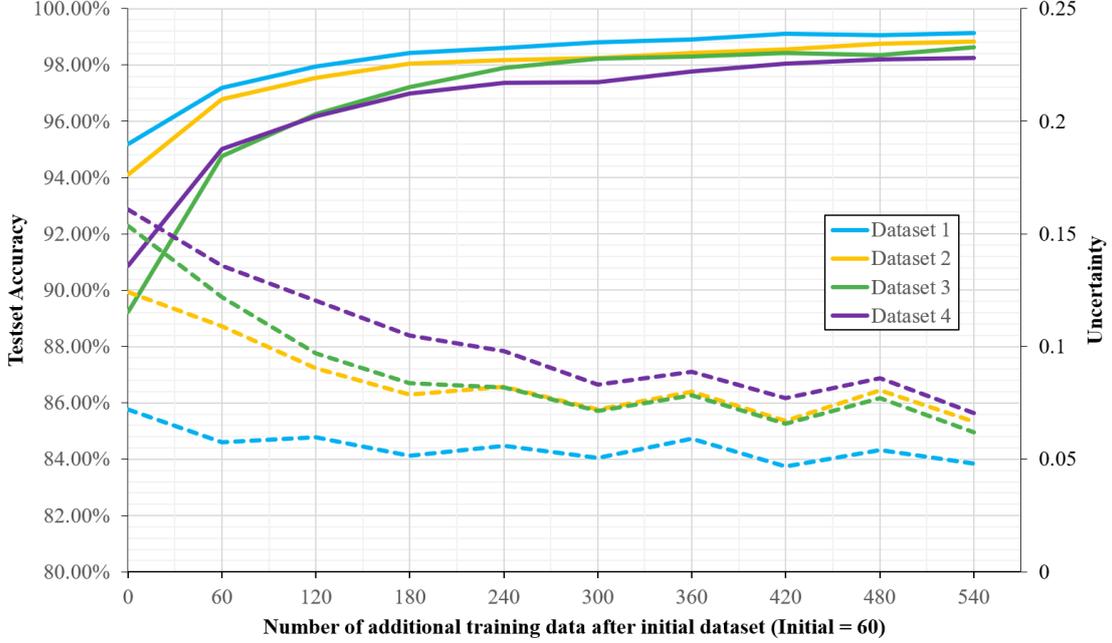

Fig. 6. Results of Experiment 1

### 4.4 Experiment 2-1
**Dataset preparation**

The set of Experiments 2 aims to demonstrate the effectiveness of the proposed uncertainty-guided active survey framework. In Experiment 2-1, we construct a synthetic dataset that contains 10,000 trip cases from various origins to 20 destinations. Each destination contains 500 trip cases with different levels of diversity (**Fig. 5b**). Similar to Experiment 1, the travel time and cost for biking, subway and driving serve as input features, and the generated travel mode choice results among bike, subway, and car serve as the ground truths (see **Appendix B** for details). These input features and ground truths constitute all the data pairs (trip cases) used for model training, validation, and testing. We split 30% of the trip cases as the test set, with the same number of trip cases for each destination. From the remaining data, we randomly select 200 cases as the validation set, and 2 trip cases for each destination (40 cases in total) as the initial training set, and other trip cases remain in the data pool for uncertainty quantification and additional data acquisition.

**Experiment settings**

We conduct the experiment following the framework proposed in Section 4.1 and **Fig. 4** and adopted simple random sampling (Method 1) and uncertainty-guided sampling strategy (Method 3) respectively to acquire additional training data to develop BTMP models. In each framework iteration, we use MC Dropout to quantify the average uncertainty of the model predictions on trips to each destination following the process mentioned in Section 3.2 and **Fig. 1**, and iteratively acquire 10 trip cases each time as additional training data using the two methods. We then compare the BTMP model's predictive performance and the amount of required training data when using different sampling strategies of Method 1 and 3. The experiment is repeated five times and the results are averaged.

**Experiment results**

As shown in **Fig. 7**, when adopting Method 3 (the uncertainty-guided sampling strategy) as additional training data acquisition method, the gradually trained BTMP model consistently outperforms the one trained using Method 1 (simple random sampling strategy). BTMP model trained using Method 3 always achieves better accuracy faster than Method 1. For instance, to achieve an accuracy of 98.0%, Method 3 only requires 170 additional training data, while Method 1 requires 270, which is 58.8% more samples than Method 3.

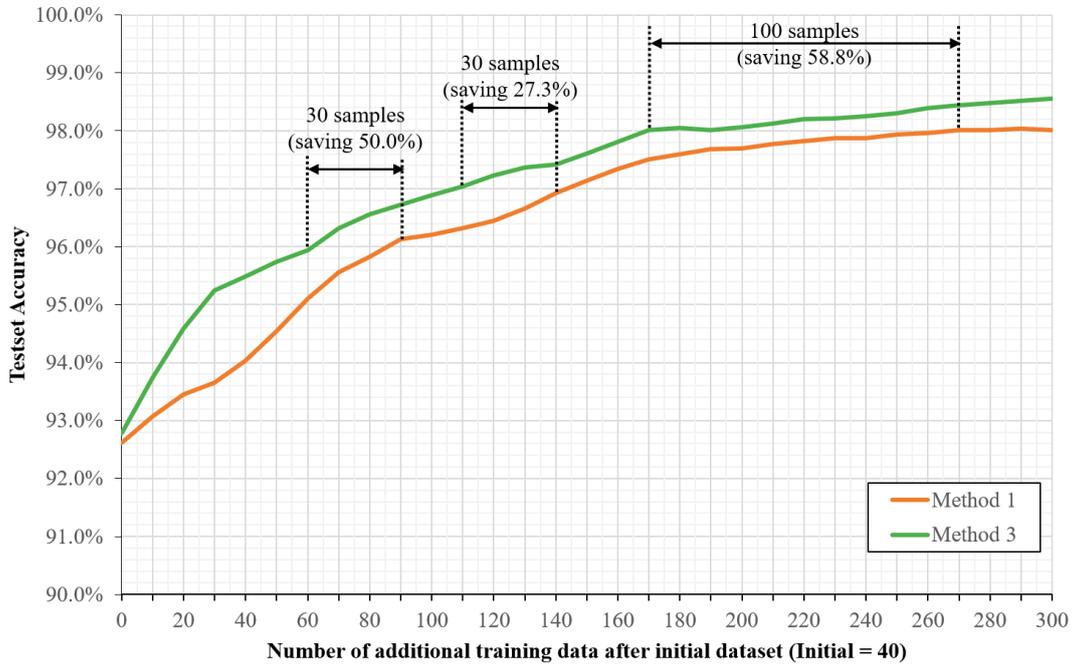

Fig. 7. Results of Experiment 2-1

**4.6 Experiment 2-2**
**Dataset preparation**

In Experiment 2-2, we apply the the alternative framework present in Section 4.1 and **Fig.4** to real-world case to further demonstrate the effectiveness of the proposed uncertainty-guided active survey framework in saving BTMP model training samples and reducing survey costs. In May 2023, we conducted a large-scale travel survey at a typical university in Beijing, China, obtaining 7367 valid travel mode choice records covering all grades of undergraduate and graduate students from various departments. By analyzing the OD information provided by mobile phone data, we pre-identified 19 primary travel destinations for the students, and then inquired about their travel mode choices to these destinations in the questionnaire.

Based on the travel survey results, we construct the realworld dataset with samples composed of input features and ground truth. The input features include the availability of the mode choice alternative (e.g., bike ownership), travel time and travel cost of each mode choice alternative (calculated from online map direction APIs, following the process introduced in Section 3.3 and **Fig. 3**) and the ground truths are the students' travel mode choices among walk, bike, e-bike, bus, subway, or taxi obtained from survey. Considering that the destination with the smallest sample size has only 137 samples, in order to ensure that the sample size for each destination in the test set is the same, we randomly draw 137 * 30% = 41 samples from each destination to form the test set. Then we randomly select 2 trip cases for each destination from the remaining data (38 cases in total) as the initial training set, and other trip cases remain in the data pool for uncertainty quantification and additional data acquisition.

**Experiment settings**

We conduct the experiment following the framework proposed in Section 4.1 and **Fig. 4** using Method 1,2 and 3 as additional training data acquisition methods respectively. In each iteration, we use MC Dropout to quantify the average uncertainty of the model predictions on trips to each destination following the process mentioned in Section 3.2 and **Fig. 1**, and iteratively acquire 10 trip cases each time as additional training data through Method 1, 2 and 3. It is worth noting that in this experiment, we include Method 2 as another benchmark to facilitate sampling strategy comparison. In the questionnaire survey, we have asked each respondent to select their most frequently visited one destination out of all 19 destinations, and Method 2 randomly samples from these travel mode choice records of the most frequently visited destination reported by all respondents. This simulates revealed preference (RP) surveys, where individuals only give travel mode choice responses to their most frequently visited destination. We then compare the model's prediction performance and the amount of required training data when using different sampling strategies of Methods 1, 2 and 3. The experiment is repeated five times and the results are averaged.

**Experiment results**

As shown in **Fig. 8**, the results indicate that Method 2 always performs the worst, followed by Method 1, while Method 3 consistently outperforms the benchmark strategies. When achieving the same level of accuracy, BTMP model trained using the uncertainty-guided sampling strategy (Method 3) always requires less training data compared to the ones adopting other random sampling strategies. **Table 2** further illustrates the number of data required for different methods to train BTMP and achieve the same accuracy. The table demonstrates that employing the uncertainty-guided sampling strategy can save approximately 20% to 50% of the data compared to the benchmark methods. This advantage becomes more pronounced as the model approaches its highest accuracy. This indicates that, compared to random sampling strategies, uncertainty-guided sampling strategy (Method 3) always identifies the most informative training data that maximizes the model's generalization capability. This enables the model to achieve the desired accuracy faster with less training data, thereby reducing survey costs.

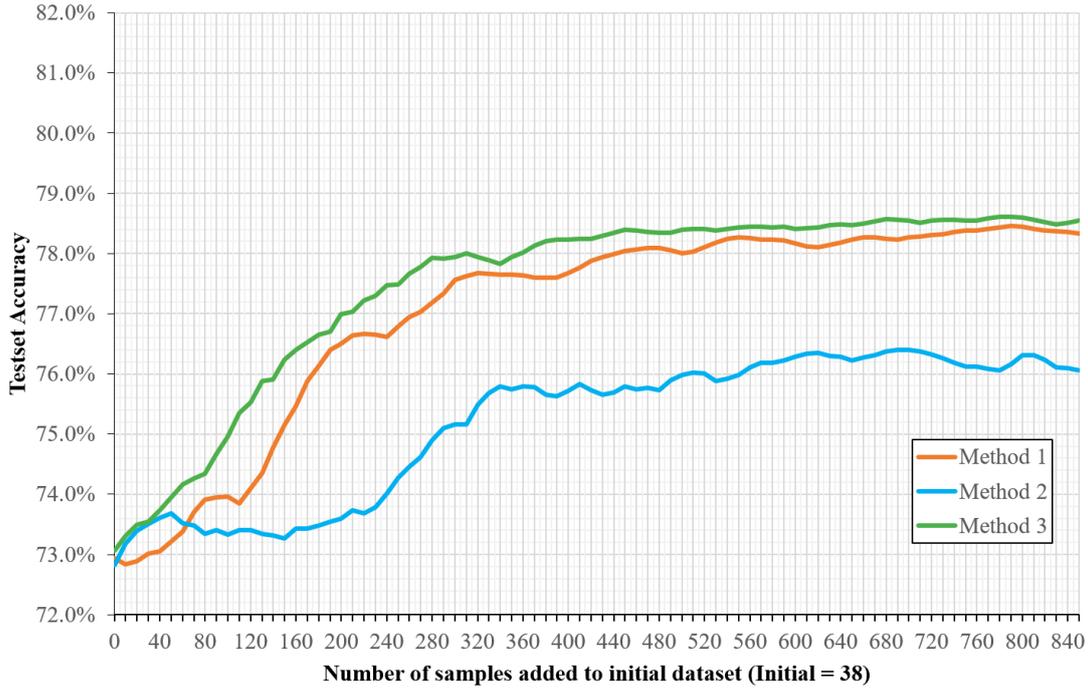

Fig. 8. Results of Experiment 2-2

Table 2
**Number of data required for different sampling strategies to reach different levels of accuracy on test set**

| Accuracy on test set | Number of Samples Required (% more data required compared to Method 3) | | |
|---|---|---|---|
| | Method 1 | Method 2 | Method 3 |
| 78.4% | 808(47.4%) | NA | 548 |
| 78% | 488(40.2%) | NA | 348 |
| 77% | 308(24.2%) | NA | 248 |
| 76% | 218(16.0%) | 548(191.5%) | 188 |
| 75% | 188(27.0%) | 328(121.6%) | 148 |

In summary, the set of Experiments 2 uses synthetic datasets and real survey data to demonstrate the effectiveness of the proposed uncertainty-guided active survey framework in saving BTMP training samples and reducing survey costs. The active survey framework dynamically formulates survey questions on destinations with high prediction uncertainty identified by gradually trained BTMP model. Through iterative collection of responses to these dynamically tailored survey questions, BTMP model gains informative training data, progressively improving its predictive performance and guiding active survey by formulating new questions with updated uncertainty information. This active survey framework supports planners in achieving better model performances with fewer questions asked, thereby reducing the survey cost.

## 5 Discussions and Conclusions

This study incorporates the concept of uncertainty in the field of explainable AI into travel mode choice modeling by proposing a BNN-based travel mode prediction model, BTMP, which is capable
17

of quantifying uncertainty of model predictions. Utilizing this capability of BTMP, this study further proposes a framework for uncertainty-guided active survey in modeling travel mode choice, which is a tailored survey process specifically designed for deep learning approaches capable of acquiring informative training data in a cost-efficient manner.

In Experiment 1, using synthetic dataset, the BTMP model shows its capability of quantifying model prediction uncertainty, which arises when the model makes predictions for scenarios beyond the scope of its training data. Experiment results also demonstrate that a dataset with higher diversity exhibits greater differences among data samples, leading to a higher probability of BTMP model encountering unseen scenarios in the test set, making the model require more training samples to build up its predictive capability with low levels of prediction uncertainty. This provides us with an opportunity to utilize the capability of BTMP to quantify uncertainty to indicate where the model needs more data to build up generalizable predictive capability.

Furthermore, the effectiveness of proposed uncertainty-guided active survey framework is tested through the performance comparison among BTMP model trained adopting uncertainty-guided additional training data acquisition method and other random sampling stratgies. Using synthetic data (Experiment 2-1) and real-world survey data (Experiment 2-2), the experiment results demonstrate that compared to random sampling strategies (Method 1 and 2), collecting responses tailored for questions with high prediction uncertainty (Method 3) significantly reduces the sample required to reach desired accuracy. Specifically, Method 1 requires approximately 20% to 50% more samples compared to Method 3 in order to reach the same desired accuracy. The uncertainty-guided sampling strategy (Method 3) always identifies the most informative training data that maximizes the model's generalization capability, further demonstrating the effectiveness of the proposed active survey framework in saving samples and reducing survey costs.

This work has made the following contributions to the research field of deep learning for travel mode choice modelling. First, the BTMP model introduced in this study innovatively integrates uncertainty quantification and explainable deep learning into the field of travel mode choice, enabling deep learning models to provide predictions while also quantifying the prediction uncertainty. In the high-risk field of transportation and urban planning, the uncertainty quantification technique proposed in this research provides guidance for decision-makers on timely human intervention when the model exhibits high prediction uncertainty, thereby avoiding potential misguidance. Additionally, the proposed active survey framework in this study enhances the efficiency of models to utilize training samples, enabling deep learning models to generalize quickly from a small amount of labeled travel mode choice data, thereby reducing the time and cost of surveys. This framework holds promise for broader applications in various types of sociological surveys and planning decision support systems, significantly contributing to the acceleration of research and decision-making efficiency in the field of transportation and urban planning.